\documentclass[review]{elsarticle}

\usepackage{lineno,hyperref}
\modulolinenumbers[5]

\journal{Pattern Recognition}

\usepackage{amsfonts,amsmath,amssymb}
\usepackage{bbm}

\newcommand{\T}{^{\ensuremath{\mathsf{T}}}} 
\providecommand{\mc}[1]{\mathcal{#1}}

\providecommand{\mbb}[1]{\mathbb{#1}}

\newcommand{\iid}{\overset{iid}{\sim}}

\usepackage{comment, subcaption}
\usepackage{caption}
\usepackage{xcolor}

\begin{document}

\title{Distance-based Positive and Unlabeled Learning for Ranking}


\author{Hayden S. Helm, Amitabh Basu, Avanti Athreya, Youngser Park, Joshua~T.~Vogelstein, Carey E. Priebe$^{*}$}
\address{
Johns Hopkins University \\ 
100 Whitehead Hall \\
3400 North Charles Street \\
Baltimore, MD 21218 USA \\
$^{*} $corresponding author; email: cep@jhu.edu
}

\author{Michael Winding, Marta Zlatic, Albert Cardona}
\address{University of Cambridge \\ Cambridge CB2 1TN UK}

\author{Patrick Bourke, Jonathan Larson, Marah Abdin, Piali Choudhury, Weiwei~Yang, Christopher W. White}
\address{Microsoft Research \\ Redmond, WA 98052 USA}

\begin{abstract}
Learning to rank -- producing a ranked list of items specific to a query and with respect to a set of supervisory items -- is a problem of general interest. The setting we consider is one in which no analytic description of what constitutes a good ranking is available. Instead, we have a collection of representations and supervisory information consisting of a (target item, interesting items set) pair. We demonstrate analytically, in simulation, and in real data examples that learning to rank via combining representations using an integer linear program is effective when the supervision is as light as "these few items are similar to your item of interest." While this nomination task is quite general, for specificity we present our methodology from the perspective of vertex nomination in graphs. The methodology described herein is model agnostic.
\end{abstract}

\maketitle

\section{Introduction}

Given a query, a collection of items, and supervisory information, producing a ranked list relative to the query is of general interest. 
In particular, learning to rank \citep{liu2009learning} and algorithms from related problem settings \citep{conte2004thirty} have been used to improve popular search engines and recommender systems and, impressively, aid in the identification of human traffickers \citep{fishkind2015vertex}.

When learning to rank, for each training query researchers typically have access to (feature vector, ordinal) pairs that are used to learn an ordinal regressor via fitting a model under a set of probabilistic assumptions \citep{robertson2009probabilistic} or via deep learning techniques \citep{severyn2015learning} that generalize to ranking items for never-before-seen queries. A query is an element of a set of possible queries $ Q $ and the items-to-be-ranked are elements of a nomination set $ \mc{N} $.

In this paper we consider the setting in which, for a given query, we know the dissimilarity (from multiple perspectives) between it and a set of items to be ranked. We are also given a set of items known to be similar to the query (positive examples). 
We make no model assumptions.
Our goal is to leverage the knowledge of the items known to be similar to the query to produce a new dissimilarity tailored to the query. The new dissimilarity, which in this paper is exactly a convex combination of the different dissimilarities, is then useful for nominating items unknown to be similar to the query. This approach to learning a ranking scheme can gainfully be seen as combining representations to improve inference
\citep{vogelstein2020general}.

In the language of graphs:
with respect to a specific vertex of interest $ v^{*} $,
and given a set $S$ of important vertices and a collection of dissimilarity measures,
we solve an integer linear program that weights those dissimilarities so that the set of points in $S$ are ranked minimally;
we then use that learned weighted representation to rank all other vertices.
In short, we infer an entire ranking from light supervision in the form of the set $S$.

Our setting is closely related to positive and unlabeled (PU) learning \citep{Bekker_2020} in that supervision is of the form of positive examples and that unlabeled data are assumed to be neither positive nor negative. The setting herein differs from canonical PU learning in two major ways. First, in PU learning the inference task is to label unknown-to-be-positive objects as positive. For this task, standard classification algorithms such as Naive Bayes and Support Vector Machines can be modified to produce a decision function under certain sampling assumptions \citep{mordelet, 10.1145/1401890.1401920}. Our task, however, is to \textit{rank} the unlabeled data. While transforming standard classifiers into ranking functions is possible, the typical transformation returns a ranking with respect to the positive conditional distribution (as opposed to with respect to the query) and so we do not study it here. Second, in canonical PU learning the data consists of feature vectors. Herein we are only given dissimilarities between objects and thus can only use standard PU learning techniques naively. We think that this change in perspective is required to learn a ranking function with respect to the query (again, as opposed to a ranking function with respect to the positive conditional distribution). Indeed, the use of PU learning in the context of recommendation systems or ranking problems is quite new, with Zhou et al. noting ``PU learning has not been extensively explored on recommender systems .." \citep{zhou2021pure} and Zhu et al. saying ``there has been no research on combining multiple PU learning algorithms for text ranking" \citep{article}. As far as we are aware, our ``dissimilarity-based" PU learning for ranking is a novel setting and is to PU learning as dissimilarity-based pattern recognition \citep{10.5555/1197035} is to classical pattern recognition \citep{devroye2013probabilistic}.

Related to our work, \citep{fagin1999combining} and the corresponding literature \citep{fagin2002combining} deal with a multi-media system (a system consisting of many subsystems) where the multi-media query $v^{**}$ is a boolean function of different $v^*_{1}$, $v^{*}_{2}, .. v^{*}_{m}$ 
with $v^{*}_{i}$ a query corresponding to subsystem $i$. Each subsystem is assumed to be able produce a ranked list for its corresponding query using fuzzy logic. In the context of our discussion, it is fruitful to think of these ranked lists as coming from different (marginal of subsystem $i$) representations of $v^{**}$. Fagin proposes an optimal algorithm for combining these representations under a set of assumptions --  most notably that the boolean function that combines the subsystem queries is known and that the subsystems are independent. These two assumptions, in particular, imply that supervisory information is not necessary. In our setting we make no assumptions on the structure of the query nor on the relationship between the representations and hence rely on supervisory information to combine representations.

Our work is also related to the set expansion literature \citep{10.1007/978-3-319-71249-9_18, 10.1145/1645953.1645961, 10.1145/1963405.1963467}. Most similar to the ideas discussed in this paper here is the algorithm SetExpan proposed by Shen and Wu et al. \citep{10.1007/978-3-319-71249-9_18} where they combine information from a set of ranked lists from different contexts to iteratively add to their set of important items. There are three main differences between SetExpan and our approach. The first is that for SetExpan the resulting ranked list is relevant to the entire $ S \cup \{v^{*}\} $ set and not just the vertex of interest $ v^{*} $. Herein we do not assume symmetry in the relationship between $ v^{*} $ and the elements of $ S^{*} $. The second is that SetExpan uses the elements of $ S \cup \{v^{*}\} $ to find a subset of the representations to use in the final ranked list. We, on the other hand, our approach uses the elements of $ S $ as supervision to a learning problem that optimizes a linear combination of the original representations. Lastly, conditioned on the selected subset of representations, SetExpan uses a simple average to combine the representations. Our approach uses a learned convex combination instead.

While our set up is quite general, we study it through the lens of vertex nomination \citep{JMLR:v20:18-048}.

%
%



\section{Problem Description: Vertex Nomination}

In the single graph vertex nomination problem 
\citep{JMLR:v20:18-048}
we are given a graph $ G = (V, E) $ and a single vertex of interest $ v^{*} \in V $,
and the task is to find other interesting vertices.
The vertex set $ V $ can be taken to be 
$V = [n] = \{1, \hdots, n\} $
and the edge set $ E \subset {{V}\choose{2}}$ is a subset 
of all possible vertex pairs $\{i,j\}$ with $ i, j \in [n] $.
The objective of vertex nomination is to return a ranked list of the candidate vertices $ V \setminus \{v^{*}\} $ such that ``interesting'' vertices
-- vertices ``similar'' to $ v^{*} $ --
are ranked high in the nomination list. Note that in vertex nomination the query set $ Q $ is $ \{v^{*}\} $ and the nomination set $ \mc{N} $ is $ V \setminus \{v^{*}\} $.


Vertex nomination is a special case of (typically) unsupervised problems addressed by recommender systems \citep{bobadilla2013recommender} where it is assumed that {\it "[i]nformation relevant to the task is encoded in both the structure of the graph and the attributes on the edges"} \citep{coppersmith2014vertex}. There have been numerous approaches to vertex nomination proposed in recent years \citep{marchette2011vertex, coppersmith2012vertex, sun2012comparison, suwan2015bayesian, fishkind2015vertex, agterberg2019vertex, yoder2020vertex} with each illustrating success in sometimes adversarial settings. 

Notably, none of these proposed nomination schemes is universally consistent. Recall that a universally consistent decision rule is one where the limiting performance of the decision rule is Bayes optimal for every possible distribution of the data. In the classification setting, for example, the famed $ K $ Nearest Neighbor rule \citep{fix1951discriminatory}, with appropriate restrictions on $K$ growing with training set size, is in a class of decision rules known to be universally consistent, \citep[Chapters~5,6]{devroye2013probabilistic}. In their foundational paper on the theoretical framework of vertex nomination, Lyzinski et al.\ show that there does not exist a universally consistent vertex nomination scheme \citep{JMLR:v20:18-048}. Their paper complements other theoretical \citep{peel2017ground} and empirical \citep{Priebe5995} results on the limitations of machine learning for popular unsupervised learning problems on graphs. The successes reported in \citep{marchette2011vertex, coppersmith2012vertex, sun2012comparison, suwan2015bayesian, fishkind2015vertex, agterberg2019vertex, yoder2020vertex} were all with respect to some application-specific notion of similarity/interestingness. 

In this paper,
in contrast to being told what is meant by similarity,
we consider the setting in which, 
in addition to $ G $ and $ v^{*} $, 
we are given a set of vertices $ S \subset (V \setminus \{v^{*}\}) $ explicitly known to be similar to $ v^{*} $
from which we are to 
{\it learn} 
a ranking scheme
specific to the task at hand.
In particular, we develop a nomination scheme $ f $ that takes as input $(G,v^*,S)$ -- a graph, a vertex of interest, and a set of vertices known to be similar to the vertex of interest --
and outputs a function that maps each vertex
not equal to $ v^{*} $ to an element of the set $ [n-1] $. 
(We ignore the possibility of ranking ties for expediency;
see Appendix B.1 of \citep{JMLR:v20:18-048} for a discussion.)

We consider $S^* \subset (V \setminus \{v^*\})$
to be the collection of vertices that are truly similar to $v^*$;
thus the given supervisory set $S \subset S^*$
and the set $(S^* \setminus S)$,
representing an unknown truth,
consists of vertices that we actually want to identify as interesting by placing them highly in the nomination list.
Letting $ \mc{H} = \{h: V \setminus \{v^{*}\} \to [n-1]\} $ be the set of functions, or rankers, that map a vertex to an element of $ [n-1] $ and $ \mc{G}_{V} $ be the set of graphs with vertex set $V$, a nomination scheme is a mapping $ f: \mc{G}_{V} \times V \times 2^{V} \to \mc{H} $. Our goal is to use an $ f $ such that $ f(G, v^{*}, S) = h(\cdot) $ outputs small values for elements of $S^* $.

We refer to the set $ C = (V \setminus (\{v^*\} \cup S)) $ as the candidate set and note that $ (S^{*} \setminus S) \subset C $.
For evaluation purposes it is convenient to consider the set of rankers that map from $ C $ to the nomination range $ R = [n-1-|S|] $. When there is a possibility of confusion we denote such rankers as $ h_{C} $, with $ \mc{H}_{C} = \{h_{C}: C \to R\} $. For every ranker $ h \in \mc{H} $ there is a ranker $ h_{C} \in \mc{H}_{C} $ such that $ h_{C}(v) $ is equal to $ h(v) $ minus the number of elements of $ S $ ranked higher than $ v $ in the nomination list induced by $ h $ for all $ v \in C $.



\section*{Methods}

\subsection*{Natural Nomination, Given a Dissimilarity}


Recall that $ h $ is a mapping
from the set of vertices minus $ v^{*} $ to the set of ranks $ \{1, \hdots, n - 1\} $.

Given a dissimilarity measure $ d: V \times V \to \mbb{R} $,
a natural ranking function to consider is one that, given a vertex $v \neq v^{*} $, returns the rank 
of the real number $ d(v^*,v) $ amongst the collection $\{d(v^*,v')\}_{v' \neq v^{*}}$.
That is, the vertex $ v \in (V \setminus \{v^{*}\}) $ that minimizes $ d(v^{*}, v) $
is mapped to 1 -- the top of the nomination list -- and the vertex farthest from the vertex of interest is mapped to $ n - 1 $. 
We let $ h^d $ denote this mapping from the vertex set to the set $ [n-1] $ for dissimilarity $ d $.

We emphasize that throughout our discussion $ d $ need not satisfy the symmetry, triangle inequality or non-negativity requirements of a metric.






\subsection*{An Integer Linear Program} \label{subsec:IP}

We present an optimization problem whose solution is useful for learning to rank in general and supervised vertex nomination in particular. 
Let $\{v_1, \ldots, v_n\}$ be a finite set of items.
Without loss of generality we identify $v_1=v^*$.
We have a collection of $J$ distinct dissimilarity measures $ \{d^{1}, d^{2}, \hdots, d^{J}\} $ and have knowledge of the dissimilarity between $v_1$ and $v_i$, $i=2, \ldots, n$ for each of these measures. We use $d^j(v_{1}, v_{i}) $ to denote the dissimilarity between $v_1$ and $v_i$ in the $j$-th dissimilarity, $j=1, \ldots, J$. We are given a set $S \subset \{v_2, \ldots, v_n\}$ that we want to rank as high as possible by choosing an appropriate weighted combination of the $J$ dissimilarities. More precisely, we wish to select a set of weights $\alpha_1, \ldots, \alpha_J \geq 0$ such that when the elements $v_2, \ldots, v_n$ are ranked according to the dissimilarity $\alpha_1d^1(v_{1}, v_{i}) + \ldots + \alpha_Jd^J(v_{1}, v_{i})$ (for $i=2, \ldots, n$), the elements of $S$ are as close to the top of the ranked list as possible. 

Formally, for a given tuple of weights $\alpha = (\alpha_1, \ldots, \alpha_J)$, let 
$ h^{\alpha}(v_{i}) $ denote the rank of $v_i$ under the dissimilarity $\alpha_1d^1(v_{1}, v_{i}) + \ldots + \alpha_Jd^J(v_{1}, v_{i})$. We wish to solve the following optimization problem: \begin{equation}\label{eq:main-prob}
\min_{\alpha \geq 0} \;\; \max_{v \in S} \;\; 
h^{\alpha}(v).
\end{equation} 

The above problem can be formulated using the framework of {\it integer linear programming} (ILP). An ILP problem is an optimization problem where one wishes to minimize/maximize a linear function of a finite set of decision variables subject to linear inequality constraints and where a subset of the variables are required to take only integer values. In our current setting, we model~\eqref{eq:main-prob} as follows.

\begin{enumerate}
\item Introduce real valued decision variables $\alpha_1, \ldots, \alpha_J$ that are constrained to be nonnegative. These are the weights we are seeking in~\eqref{eq:main-prob}. We also impose the normalization constraint that $\alpha_1 + \ldots + \alpha_J = 1$, since scaling the weights by the same positive factor yields the same solutions.
\item Introduce {\it integer} variables $x_v$ for each item $v \not\in S$. Impose the linear constraints $0 \leq x_v \leq 1$; this forces $x_v \in \{0,1\}$ in any feasible solution. These variables are to be interpreted as follows: if $x_v=0$ in any solution, then $v$ is ranked worse than every element of $S$ (under $\alpha$) and if $x_v =1$ in any solution, then $v$ is ranked better than at least one element in $S$ (under $\alpha$).
\item The linear objective function involves only the $x_v$ variables: we wish to minimize $\sum_{v \not\in S} x_v$, because this sum equals the number of elements that are ranked better than at least one element of $S$ and thus captures the objective in~\eqref{eq:main-prob}.
\item We impose the linear constraints
\begin{align*}
\sum_{j=1}^{J} \alpha_{j} d^{j}(v_{1}, s) \leq \sum_{j=1}^{J} \alpha_{j} d^{j}(v_{1}, v) + M\cdot x_v \\ 
\;\; \forall (s, v) \in (S, C) \end{align*}
where $M := \max_{i, j} d^j_i$. This constraint imposes the desired condition that for any $v \in C $, if $x_v = 0$, then $v$ should be ranked worse than every element in $S$, i.e., its dissimilarity (under $\alpha$) from $v_1$ should be greater than or equal to the dissimilarity of every element of $S$ from $v_1$. If $x_v = 1$, then since $M$ is chosen to be the maximum of all possible dissimilarities and the coefficients $\alpha_1, \ldots, \alpha_J$ sum to 1, the constraint becomes a trivial constraint that is satisfied by all such nonnegative $\alpha$ values. 
Since we are minimizing $\sum_{v\not\in S}x_v$, for any $\alpha$, if an element $v\not\in S$ is ranked worse than every element of $S$, then the optimization would set $x_v$ to $0$. Thus, in any optimal solution $\alpha^*, x^*$ to the integer program, $x^*_v = 1$ if and only if $h^{\alpha^*}(v) < h^{\alpha^*}(s)$ for some $s\in S$.
\end{enumerate}

Once the problem~\eqref{eq:main-prob} is set up as an ILP as described above, one can bring state-of-the-art algorithms and software that employ a suite of sophisticated ideas borrowed from convex geometry, number theory and algorithm design to bear upon the problem. Python implementations based on different mixed integer solvers including Gurobi \citep{gurobi}, Common Optimization INterface for Operations Research \citep{lougee2003common} and SCIP: Solving Constraint Integer Programs \citep{achterberg2009scip} are available at \texttt{https://github.com/microsoft/distPURL}.

We note that the computational complexity of the proposed ILP is a complicated function of the number of vertices, the collection of representations 
considered, and the vertices $S$ known to be similar to the vertex of interest $ v^{*} $. 
In general, when the worst ranking element of $ S $ is ranked sufficiently poorly then the ILP is computationally burdensome. This issue is compounded when at least one element of $S$ is ranked poorly.

Further, the objective function \eqref{eq:main-prob} is but one natural choice for the problem of using the elements of $ S $ to learn a useful ranked list. Others include the $ \min $ of the average rank of elements of $ S $ and the $ \max $ of the average reciprocal rank of elements of $ S $. Understanding the performance and computational consequences of different objective functions is a promising route for future research.

\subsection*{The Solution Nomination List}

The $\alpha^*$ given by the solution to the integer program induces a dissimilarity $d^* = \sum_{j=1}^k \alpha^*_j d^j$, and the resultant $h^{d^*}$ provides an nomination list learned for $v^*$ from $ S$. We note that the ranker induced by $ d^* $ is not necessarily unique and is an element of the set $ \mc{H}^{*} = \{h: h \text{ minimizes } \eqref{eq:main-prob}\} $ whose constituent deciders map the elements of $ S $ close to the top of a nomination list.

\subsection*{Comparing two rankers}




In the simulations and real data experiments below we compare nomination schemes using Mean Reciprocal Rank (MRR). MRR is one of many measures commonly used in information retrieval to evaluate a nomination list for a given set of objects \citep{radev2002evaluating}. Let $ h $ be a ranker and $ \mc{N}' $ be a subset of the nomination objects $ \mc{N} $. The MRR of $ h $ for $ \mc{N}' $ is the average of the multiplicative inverse (or reciprocal) of the $ h(s) $. That is,
\begin{align*}
    \text{MRR}(h, \mc{N}') = \frac{1}{|\mc{N}'|}\sum_{s \in \mc{N}'} \frac{1}{h(s)}.
\end{align*} For a given $ \mc{N}' $ and two rankers $ h $, $ h' $, the ranker $ h $ is preferred to the ranker $ h' $ for $ \mc{N}' $ if $\text{MRR}(h, \mc{N}') > \text{MRR}(h', \mc{N}')$. In our experiments, $ \mc{N}' = S^{*} \setminus S $.

\section*{A Generative Model Example}


Latent space network models \citep{hoff2002latent} are random graph models where each vertex has associated with it a latent vector and the probability of an edge between two vertices is determined by a function of two vectors, typically called a \textit{kernel}. One such latent space model is the Random Dot Product Graph (RDPG) where the kernel function is the inner product \citep{athreya2017statistical}.

We consider latent positions $ X_{1}, .., X_{n} \iid F $ on $\mbb{R}^m$ 
associated with the vertices
$v_{1}, .., v_{n}$
where the distribution $ F $ is such that  
$0 \le \langle x,y \rangle \le 1 $
for all $x,y$ in the support.
Let $ X $ denote the $n \times m$ matrix with the $X_{i}$'s as rows. That is,
\begin{align*}
X = \begin{bmatrix} X_{1}^{\T} \\ \vdots \\ X_{n}^{\T} \end{bmatrix}.
\end{align*}
Then $ P = X X^{\T}$ is the $n \times n$ RDPG connectivity probability matrix.
Let $ T^{1}, \hdots, T^{J} $ be different embedding functions; that is, each $T^j: \mathcal{M}_n \to (\mbb{R}^{m_{j}})^n$ takes as input an $ n \times n $ matrix and outputs $ n $ points in $\mbb{R}^{m_{j}}$. For example, the adjacency spectral embedding of $ P = U_{P} \Sigma_{P} U_{P}^{\T} $ is an embedding with $ T_{ASE}(P) = U_{P}|\Sigma_{P}|^{1/2} $. We let $ T(P)_{i} $ denote the representation of node $ i $ resulting from the transformation $ T $. 
Then $ T_{ASE}(P)_{i} = X_{i} $ (up to an orthogonal transformation).
Further suppose that with each embedding function comes a dissimilarity $ d^{j}: \mbb{R}^{m_{j}} \times \mbb{R}^{m_{j}} \to [0, \infty) $. This induces a $v_1$-specific ``personal'' dissimilarity matrix
\begin{align*}
\Delta_{v_{1}} = 
    \begin{bmatrix} 
        d^{1}(T^{1}(P)_{1}, T^{1}(P)_{1}) & .. & d^{J}(T^{J}(P)_{1}, T^{J}(P)_{1}) \\
        \vdots & \ddots & \vdots \\
        d^{1}(T^{1}(P)_{1}, T^{1}(P)_{n}) & .. & d^{J}(T^{J}(P)_{1}, T^{J}(P)_{n})
    \end{bmatrix}
\end{align*} containing the dissimilarities from the (representation of the latent position for) the vertex of interest 
(without loss of generality, we are letting the vertex of interest be index 1: $v^*=v_1$) 
to every other vertex for every transformation in terms of its induced dissimilarity. Recall that the ILP takes as input a subset of vertices $S$ and a personal dissimilarity matrix $\Delta_{v^{*}}$.

As above, our goal is to construct a dissimilarity $ d' = \sum \alpha_{j} d^{j} $ with $ \sum \alpha_{j} = 1 $, $ \alpha_{j} \ge 0 $  such that $ d'(v_{1}, v_{s^{*}}) $ is "small"
for the elements of $ (S^{*} \setminus S) $, the vertices truly, but unknown to be, similar to $ v^{*} $.

\subsection*{An Illustrative Analytic Example}

We illustrate the geometry of combining representations in the
RDPG model
using Laplacian Spectral Embedding (LSE) and Adjacency Spectral Embedding (ASE)
\citep{von2007tutorial, sussman2012consistent}.

Given a particular realization of the $X_i$'s, $P$ is fixed (non-random).
We consider $x_1 = [0.5, 0.5]^{T}$
and 
$x_{2}, \hdots, x_{51}$ to be realizations from the uniform distribution on the positive unit disk in $ \mathbb{R}^{2} $.
We consider two embedding functions: $ T^{1}(P) = U_{P}|\Sigma_{P}|^{1/2} $ and $ T^{2}(P) = \sqrt{n}  U_{L(P)} |\Sigma_{L(P)}|^{1/2} $, both truncated at embedding dimension $ m = 2 $, where $ L(P) = D^{-1/2} P D^{-1/2}$ with $ D_{ii} $ equal to the $ i $th row sum of $ P $. The corresponding dissimilarities are taken to be Euclidean distance. 

\begin{figure*}[ht]
    \centering
    \captionsetup[subfigure]{justification=centering}
    \includegraphics[width=\textwidth]{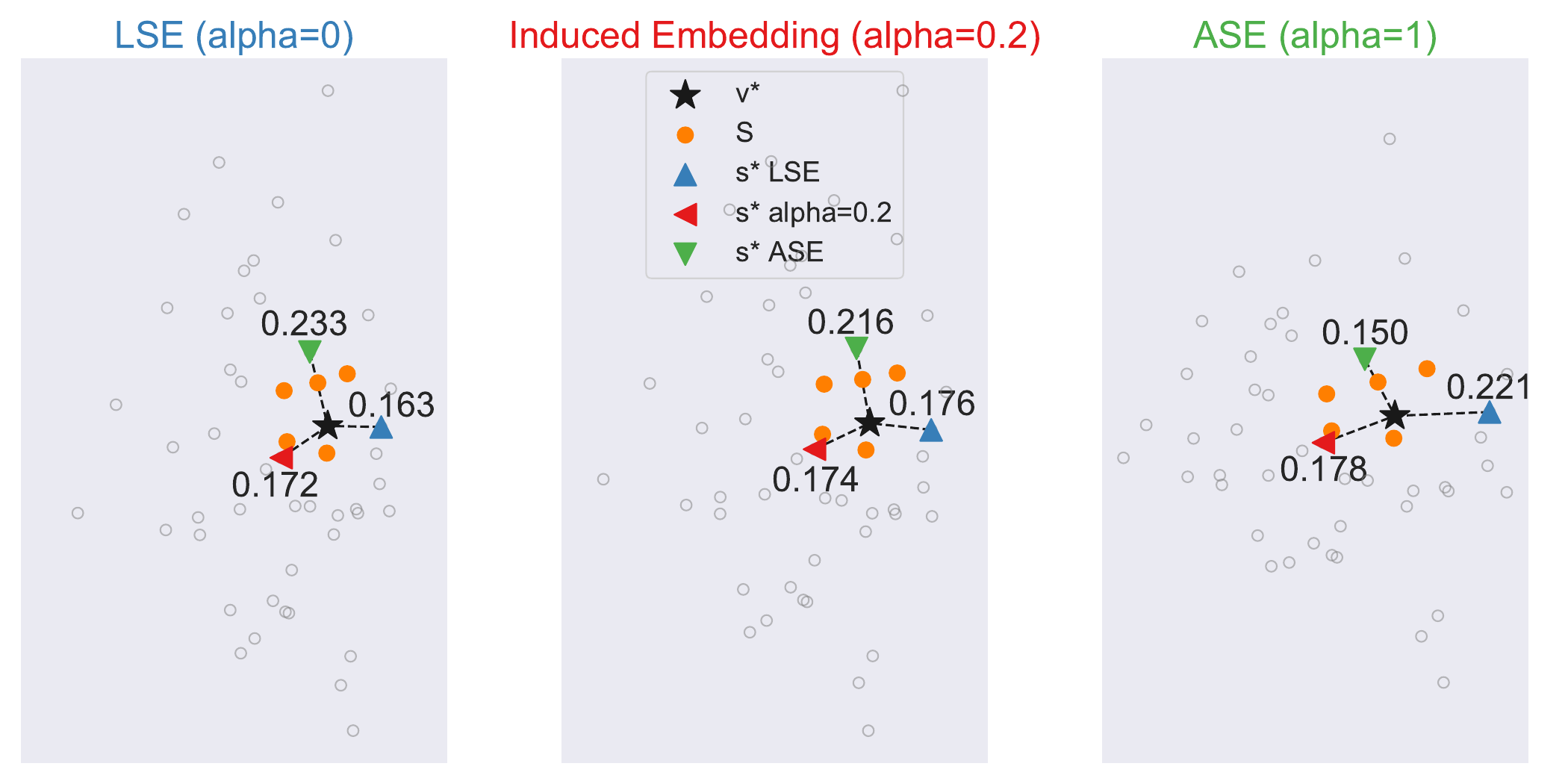}
    \caption{Ranking of interpoint distances is not preserved across vertex representations. Each panel shows an embedding of the probability matrix $ P $, with LSE on the left, the induced embedding for $ \alpha=0.2 $ in the center, and ASE on the right. In each panel, 9 vertices are highlighted: 1) $v^{*}$ -- the vertex of interest $ v^{*} $; 2-6) $S$ -- the five vertices closest to $ v^{*} $ as defined by $ d_{0.2} = 0.2 d_{ASE} + 0.8 d_{LSE} $; and 7-9) $s^{*}_{LSE}, s^{*}_{0.2} $ and $ s^{*}_{ASE} $ -- the closest vertex to $ v^{*} $ that is not an element of $ S $ as defined by $ d_{LSE} $, $ d_{0.2} $ and $ d_{ASE} $, respectively. The number near each of the three $ s^{*} $ in each panel is the Euclidean distance between it and $ v^{*} $ for that particular embedding. Note that the ordering of these distances is not preserved across panels. This implies that inference based on the ranking of the interpoint distances is not invariant to the representation of the vertices.
    }
    \label{fig:example_latent}
\end{figure*}

Figure \ref{fig:example_latent} shows that the interpoint distance rankings induced by $ T^{1} $ (ASE) and $ T^{2} $ (LSE) are not necessarily the same, thus demonstrating the basis of the ``two truths" phenomenon in spectral graph clustering \citep{Priebe5995}. Further, the interpoint distance rankings from a linear combination of the two Euclidean distances is neither equal to the rankings from ASE nor the rankings from LSE.
This indicates that the solution found by the ILP 
can produce a superior nomination list compared to either ASE or LSE alone. 

\subsection*{An Illustrative Simulation Example}

In the RDPG setting we do not observe $ P $ directly. Instead, we observe an adjacency matrix $ A $ such that $ A_{ij} \stackrel{ind}{\sim} Bernoulli(P_{ij}) $ for $ i < j $, $ A_{ji} = A_{ij} $,
and $ A_{ii} = 0 $.
Note that, save for the diagonal,
$ P = \mbb{E}(A) $. 

Revisiting our analytic example,
with geometry illustrated in
Figure \ref{fig:example_latent},
we consider the setting in which we observe $ \bar{A} = \frac{1}{k} \sum_{i=1}^{k} A_{i} $ with $ A_{1}, \hdots, A_{k} \iid Bernoulli(P) $. We define ``interestingness" based on the dissimilarity $ d_{\alpha} = \alpha d_{ASE} + (1-\alpha) d_{LSE} $ where $ d_{\{A,L\}SE} $ is Euclidean distance defined on the vertices after embedding $P$ via \{A,L\}SE. We let $ S^{*} $ be the six closest vertices to $ v^{*} $ as defined by $ d_{\alpha} $ after embedding the true but unknown probability matrix $ P $.

\begin{figure}[ht]
    \centering
    \captionsetup[subfigure]{justification=centering}       
    \begin{subfigure}{0.49\linewidth}
        \includegraphics[width=\columnwidth]{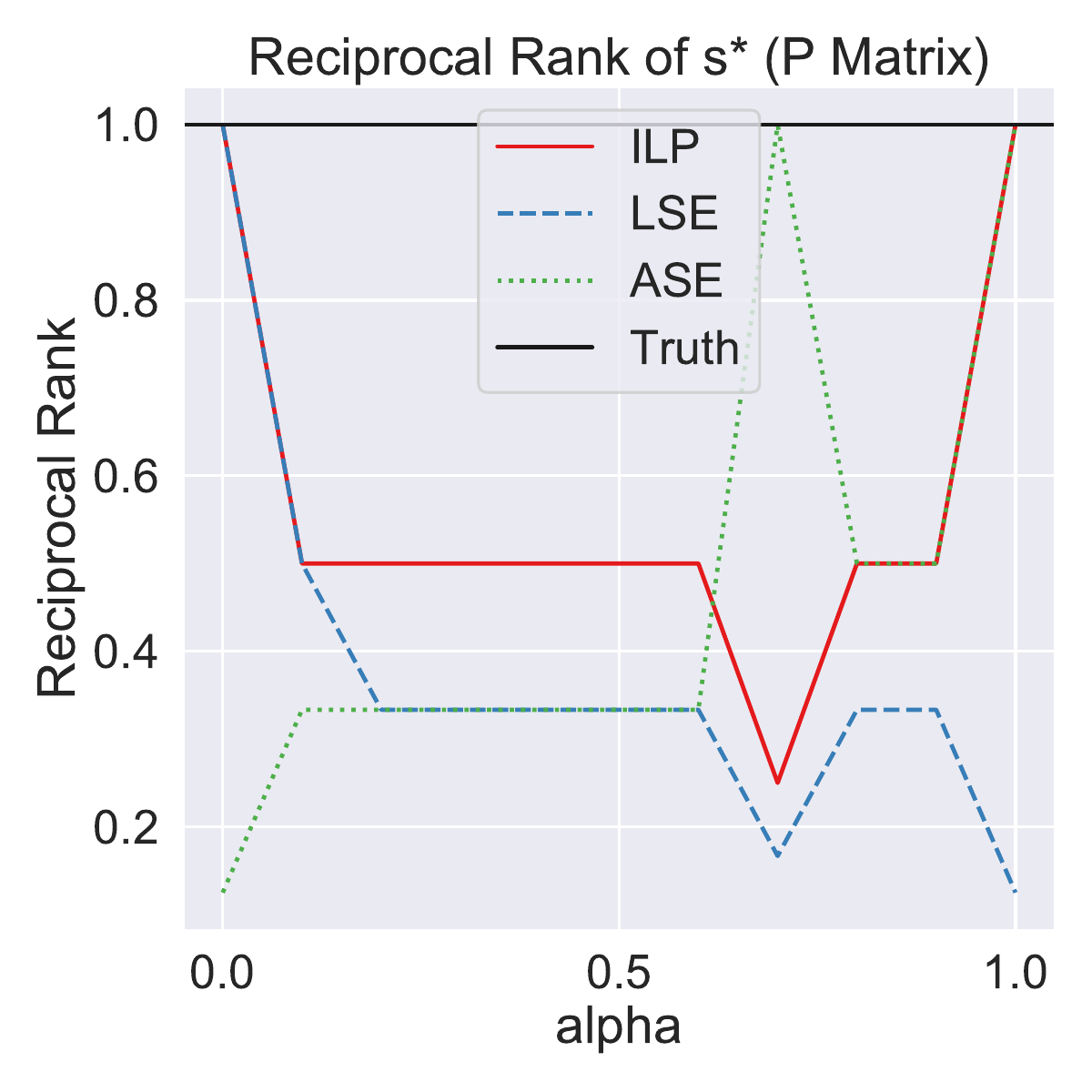}
        \caption{}
       \label{subfig:simulation_P}
    \end{subfigure}
    \begin{subfigure}{0.49\columnwidth}
       \includegraphics[width=\columnwidth]{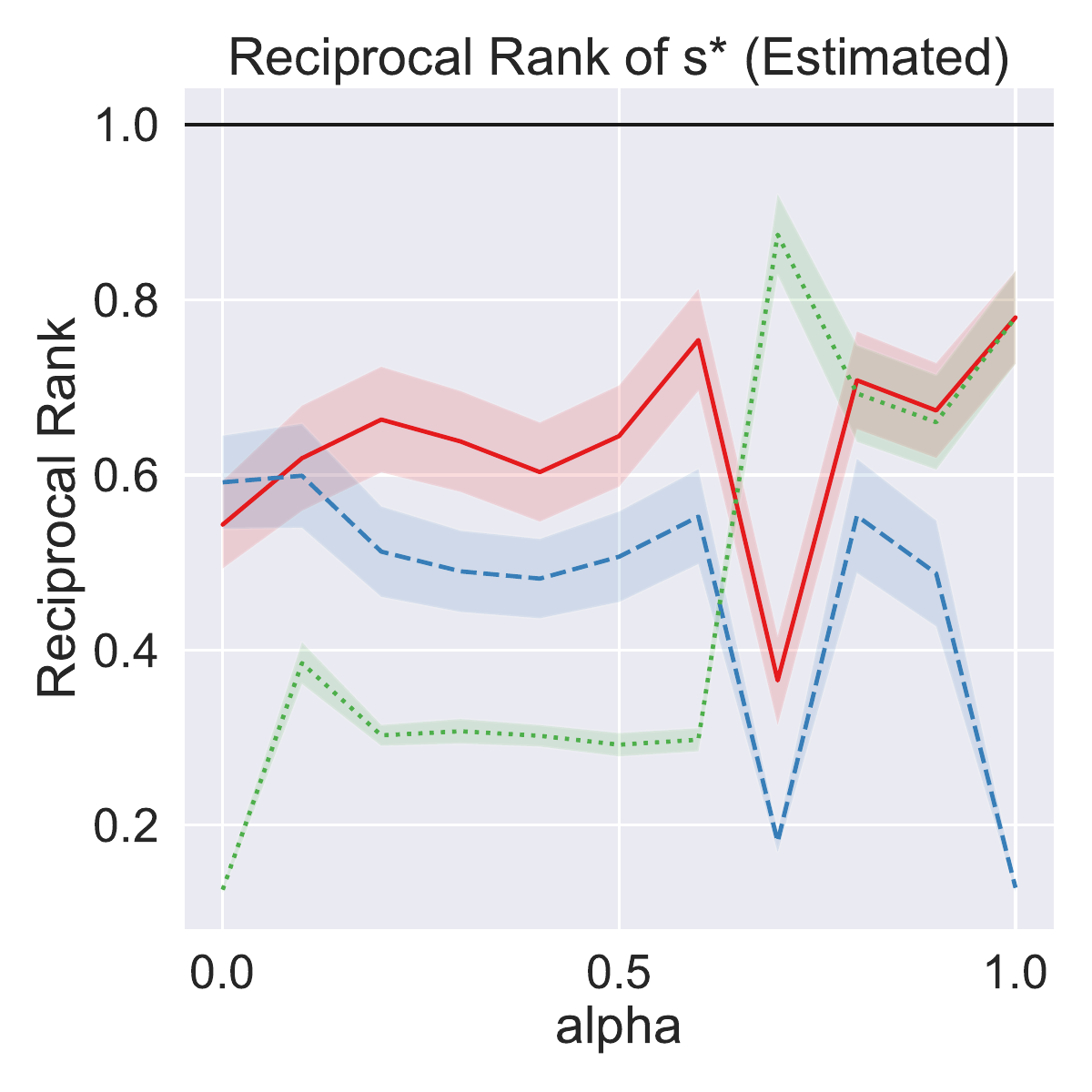}
        \caption{}
       \label{subfig:simulation_A}
    \end{subfigure}
    \caption{The proposed ILP demonstrates utility both analytically in the noiseless setting (left) and via simulation in the noisy setting (right). In both settings the definition of ``interestingness" is based on a linear combination of the distances induced by the adjacency and Laplacian spectral embeddings of the $ P $ matrix. The left panel shows the performance of the ILP and the rankers induced by ASE and LSE when the $ P $ matrix is observed. Notice that the performances of the ILP and LSE (resp.\ ASE) are the same when $\alpha$ = 0 (resp.\ 1);
    in these cases ``interestingness" coincides exactly with the distances induced by LSE and ASE. The right panel shows the performance of the same nomination scheme 
    for observed graphs instead of $ P $. 
    }
    \label{fig:example_simulation}
\end{figure}

Figure \ref{fig:example_simulation} presents the results from this simulation set up for various values of $ \alpha $ where $ S $ is the set of five closest vertices to $ v^{*} $ as defined by $ d_{\alpha} $. The rankers are evaluated based on where the sixth closest element, $ s^{*} $, as defined by $ d_{\alpha} $, is in their respective nomination lists.
The left panel shows the performance of the ILP and rankers induced by $ d_{ASE} $ and $ d_{LSE} $ when the $ P $ matrix is observed. The right panel shows the performance of the same three schemes when $ k = 1000 $. Reciprocal rank is estimated using 100 Monte Carlo simulations. Shaded regions indicate the 95\% confidence interval for the mean. The two panels demonstrate
the utility of the ILP solution for learning to nominate in both noiseless and noisy settings. We note that when $ \alpha = 0 $ ``interestingness" coincides exactly with the dissimilarity defined on the representation of the vertices after embedding via LSE and that when $ \alpha = 1 $ ``interestingness" coincides exactly with the dissimilarity defined on the representation of the vertices after embedding via ASE.

In general, it is possible that either ASE or LSE places $s^{*}$ at the top of their respective nomination lists for any given $ \alpha $. When this happens, as is the case for $\alpha=0.7$ for the latent positions described in Figure \ref{fig:example_latent}, finding a linear combination of weights that optimize \eqref{eq:main-prob} may not similarly place $s^{*}$ at the top of its corresponding nomination list. This happens, for example, when a representation that does not place $s^{*}$ at the top of its nomination list places the elements of $S$ closer to the top of its nomination list as compared to the representation that places $s^{*}$ at the top. Hence, per the objective function, the weight corresponding to the representation that does not place $s^{*}$ at the top will be larger and the performance of the ILP may suffer, as seen in Figure \ref{fig:example_simulation}.

\section*{Real Data Examples}
We consider three real data examples:
diffusion MRI connectome, search navigation, and {\it Drosophila} connectome.


\subsection*{dMRI}

\begin{figure}[t]
    \centering
    \captionsetup[subfigure]{justification=centering}
    \begin{subfigure}{0.49\columnwidth}
        \includegraphics[width=\linewidth]{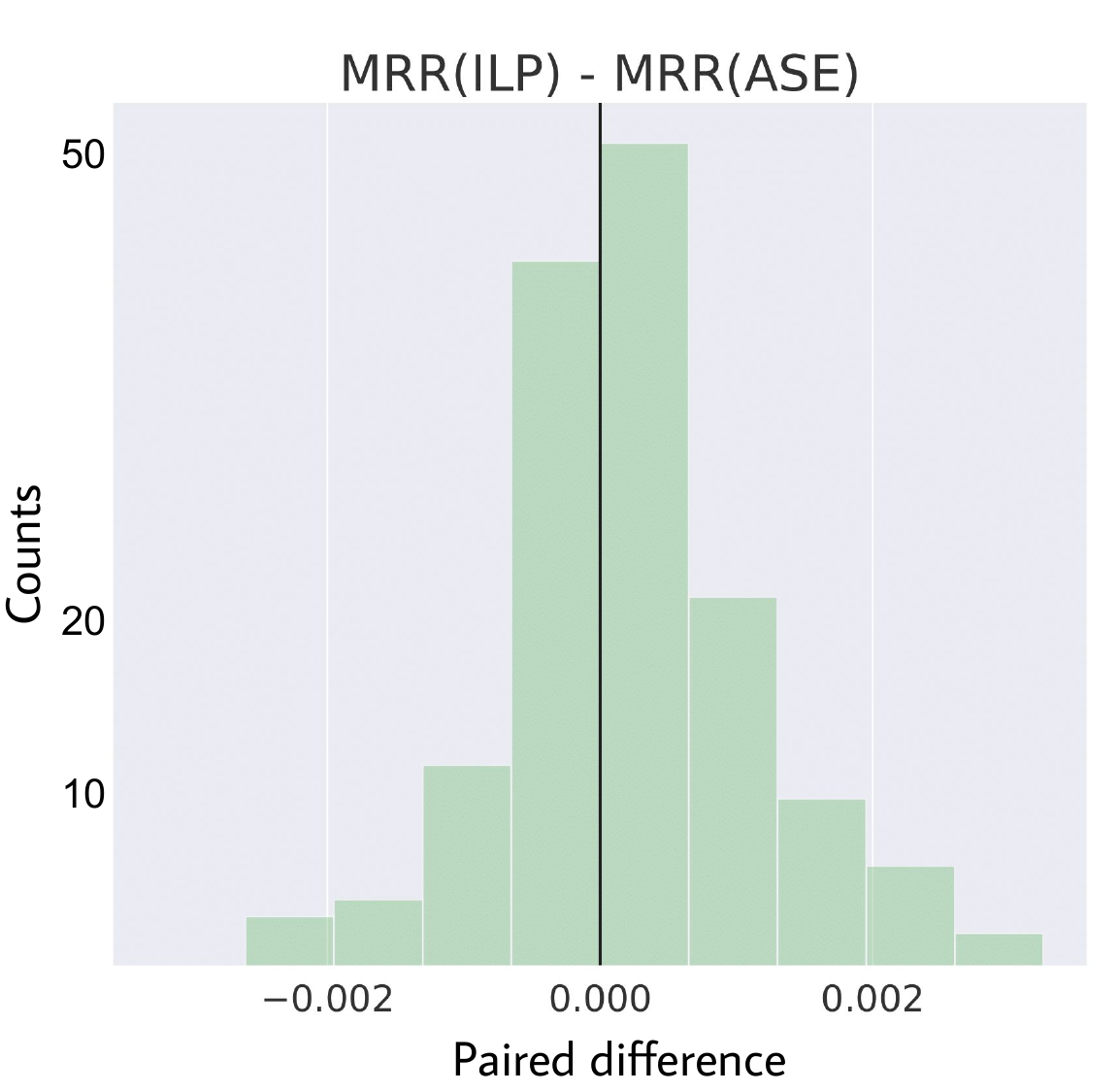}
        \caption{}
        \label{subfig:ip_minus_ase_dmri}
    \end{subfigure}
    \begin{subfigure}{0.49\columnwidth}
        \includegraphics[width=\linewidth]{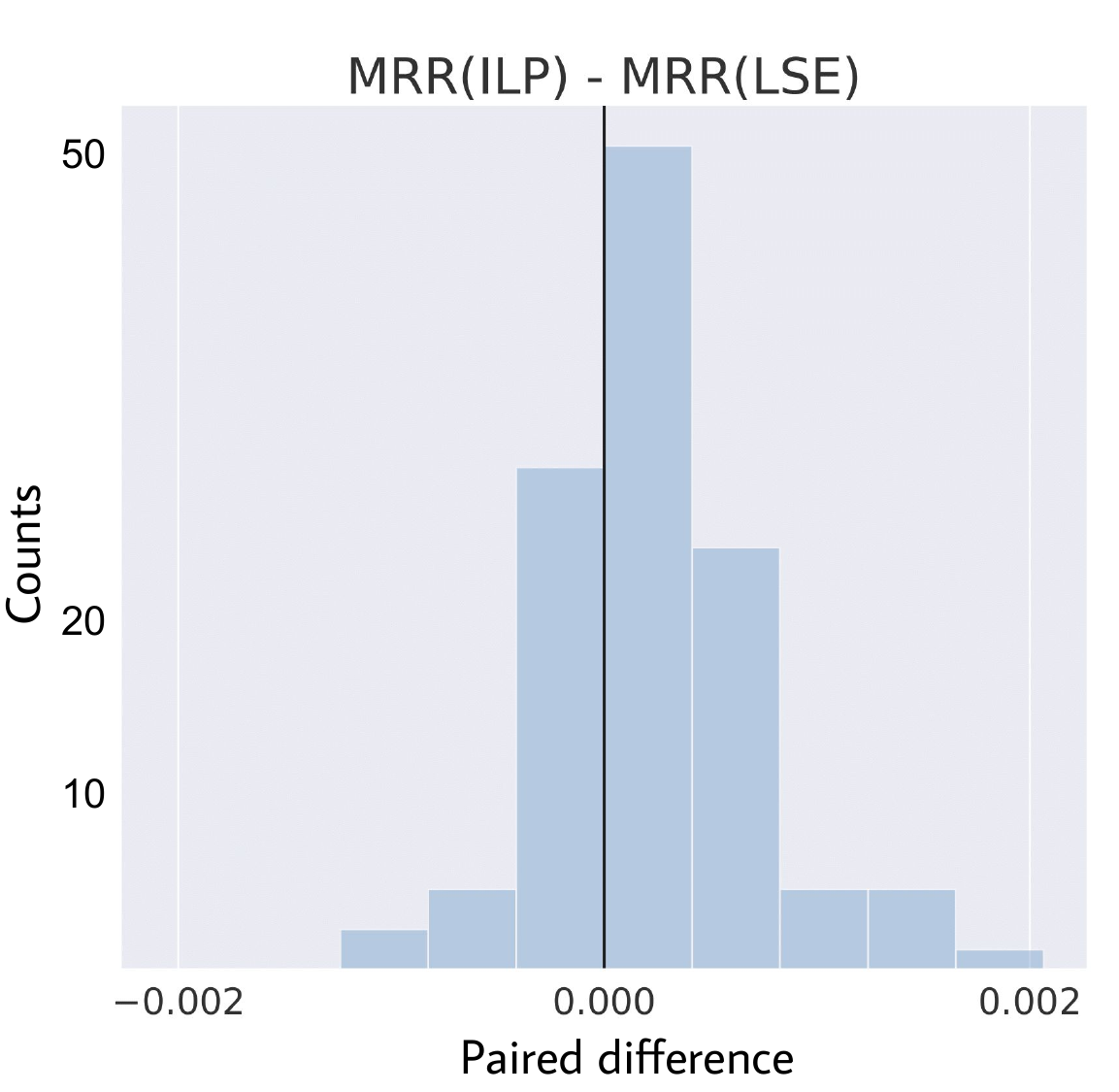}
        \caption{}
        \label{subfig:ip_minus_lse_dmri}
    \end{subfigure}
    \caption{For nominating the anterior cingulate cortex (ACC): histograms of the difference of paired Mean Reciprocal Ranks (MRR) between the nomination list induced by the Integer Linear Program (ILP) and the nomination list from ASE (a) and LSE (b). The vertex of interest $v^*$ and interesting vertex set $S$
    were sampled 
    from $V_{ACC}$. 
    MRR was calculated using the remaining elements of $V_{ACC}$. 
    Qualitatively, both distributions of differences are inclined to the positive. Quantitatively, 
    Wilcoxon's signed rank test yields p-values $ p \approx 0.0057 $ and $p < 0.00001$ for ASE and LSE, respectively,
    demonstrating a statistically significant improvement in nomination performance for the ILP solution.}
    \label{fig:ip_minus_dmri}
\end{figure}

We consider a graph $G_{dMRI}$
from a collection of connectomes estimated using a diffusion MRI-to-graph pipeline \citep{Kiar188706}. Vertices represent subregions defined via spatial proximity and edges are defined by tensor-based fiber streamlines connecting these regions. 
$ G_{dMRI} $ has $n = |V| = 40,813$ vertices and  $e = |E| = 2,224,492$ edges.

The vertices of $G_{dMRI}$ each belong to exactly one of $ 70 $ Desikan regions of the brain -- 35 anatomical regions in each of the two hemispheres \citep{desikan2006automated}. 
Furthermore, each vertex also has a designation as either gray matter or white matter. 
Thus, each vertex has a region label, a hemisphere label, and a tissue type label.

We consider
 $ J = 2 $ spectral embedding representations of $G_{dMRI}$:
ASE (embedding dimension $m=15$) 
and LSE (embedding dimension $m=46$). 
In the illustrative paper \citep{Priebe5995}, aptly titled {\it On a two-truths phenomenon in spectral graph clustering}, it is demonstrated that these two representations 
lead to two fundamentally different clusterings -- LSE best for the affinity structure associated with hemisphere (left vs.\ right) and ASE best for  the core-periphery structure associated with tissue type (gray vs.\ white).
That is, there are two truths,
and the two embeddings are each best for recovering a different truth.
In the conclusion to \citep{Priebe5995} 
the authors write
{\it ``For connectomics, this phenomenon [$\dots$] suggests that a connectivity-based parcellation based on spectral clustering should consider both LSE and ASE''}.  
The methodology developed herein allows just such an analysis.


For illustration, we consider the target brain structure to be Desikan region "anterior cingulate cortex" (ACC).
The  vertex of interest $v^*$ is chosen at random from $V_{ACC}$,
and the remainder of vertices in $V_{ACC}$ are designated as $S^*$ -- truly similar to $v^*$.
$|V_{ACC}| = 746$,
so $|S^*| = 745$.
Then $S \subset S^*$ with $|S|=50$ is randomly chosen,
leaving
$|S^* \setminus S| = 695$ truly, but unknown to be, interesting vertices out of a candidate set 
$C = V \setminus (\{v^*\} \cup S)$ with 
$|C| = n-1-|S| = 40,762$.

$V_{ACC}$
 includes vertices from both hemispheres and from both tissue types;
loosely speaking, the "truth" for $ v^{*} $, as exemplified by the characteristics of the elements of $ S $, is a combination of the original two truths. Hence, a nomination scheme that combines LSE and ASE promises superior nomination performance.

We compare the nomination schemes induced by the ASE and the LSE distances to the ILP nomination scheme that identifies an optimal linear combination of the two.

The competing nomination schemes are evaluated via MRR.
We performed this ($v^*,S^*$) sampling a total of 150 times.
The paired differences between the MRR from the ILP and ASE and between the MRR from the ILP and LSE are depicted in Figure \ref{fig:ip_minus_dmri};
positive values indicate that the ILP performs better than its competitor.
Testing for a difference in the medians between ILP and \{ASE,LSE\}
via Wilcoxon's signed rank test 
yields p-values 
$p \approx 0.0057$ for "H0: ASE as good or better than ILP" and 
$p < 0.00001$ for "H0: LSE as good or better than ILP", 
demonstrating a statistically significant improvement in nomination performance for the ILP solution.


\subsection*{Bing}

\begin{figure}[t]
    \centering
    \captionsetup[subfigure]{justification=centering}
    \begin{subfigure}{0.49\columnwidth}
        \includegraphics[width=\linewidth]{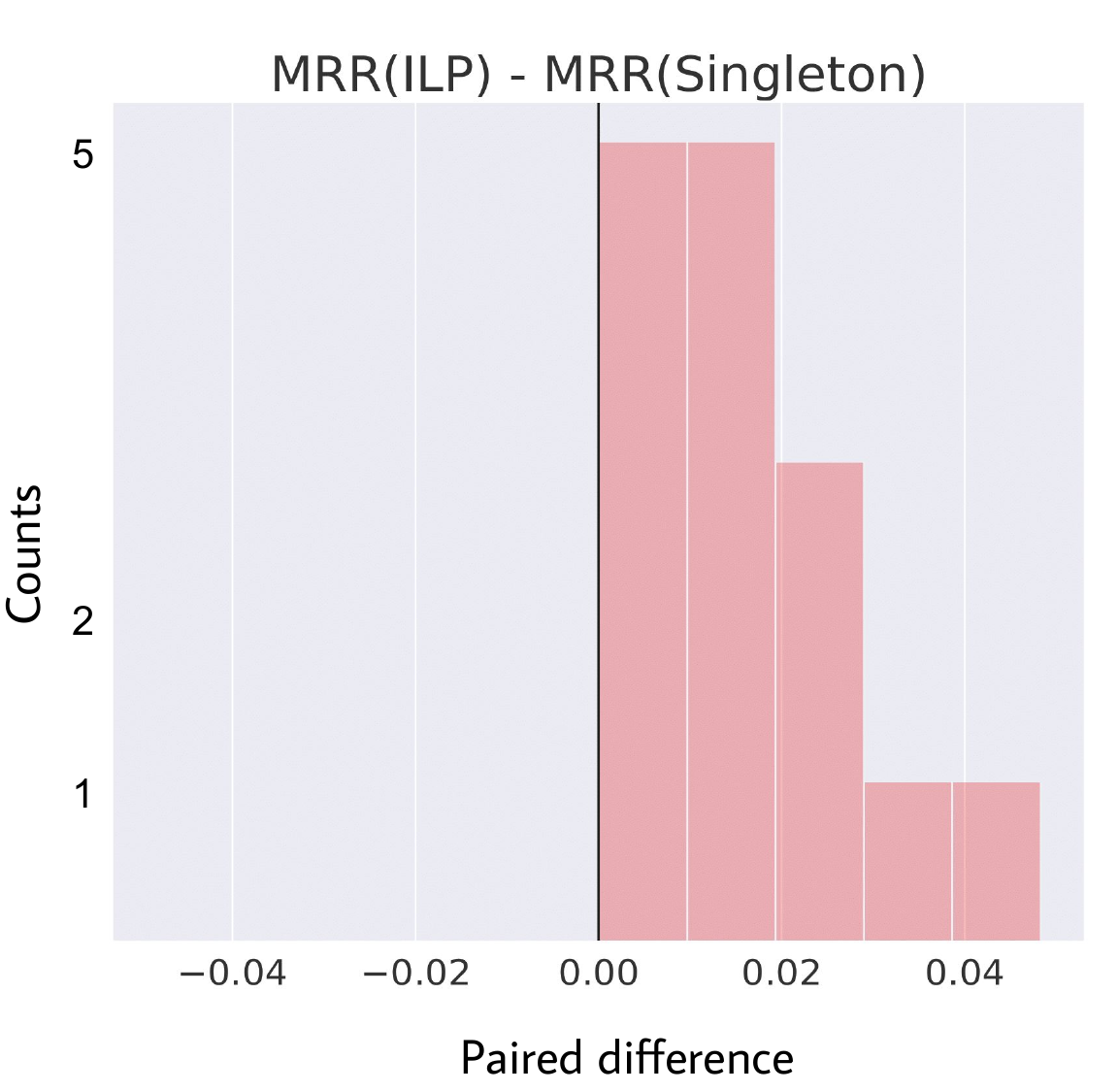}
        \caption{}
        \label{subfig:bing-mrr}
    \end{subfigure}
    \begin{subfigure}{0.49\columnwidth}
        \includegraphics[width=\linewidth]{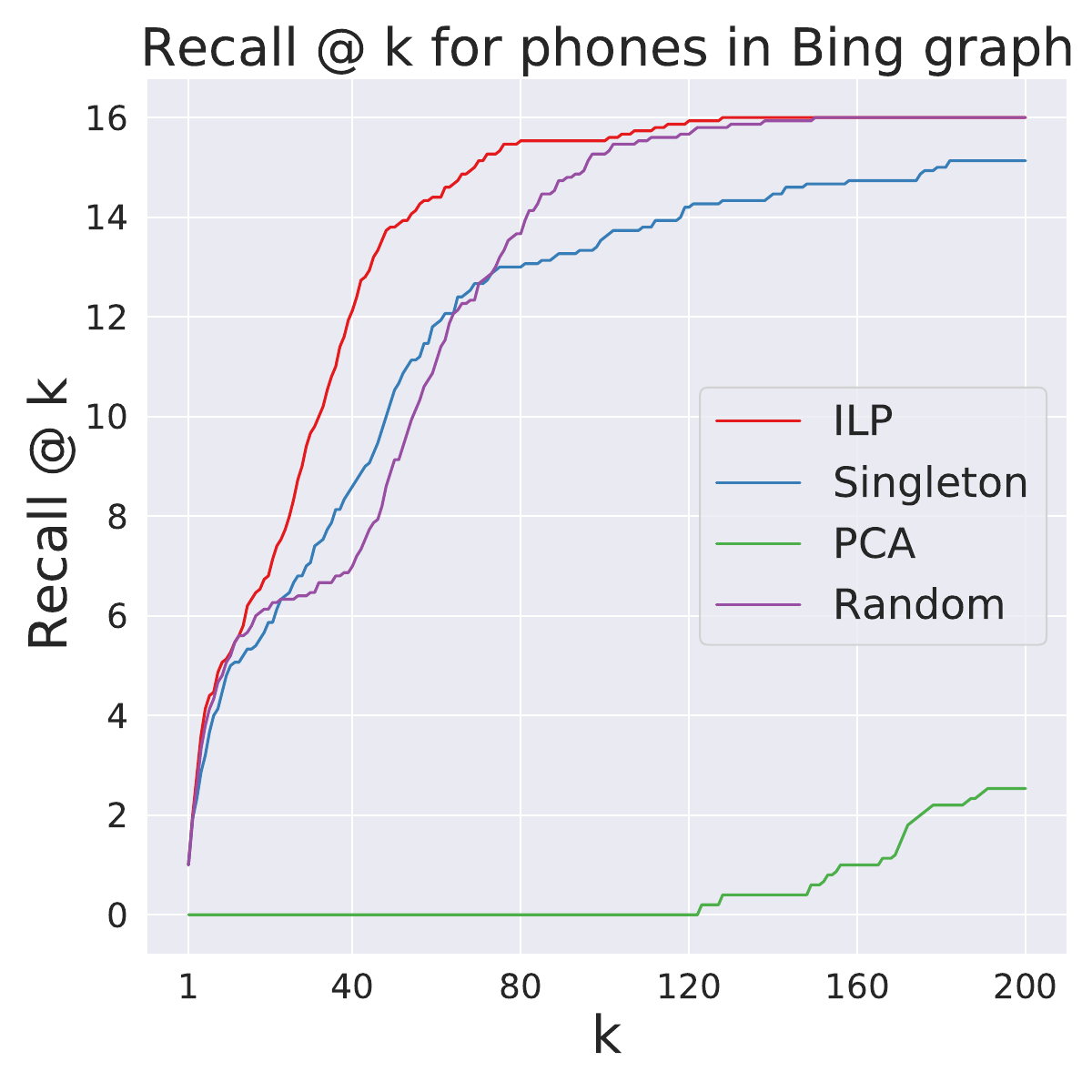}
        \caption{}
        \label{subfig:bing-recall}
    \end{subfigure}
    \caption{For nominating phones in the Bing graph, three different phones were considered to be a $ v^{*} $. Each phone has an associated $ S^{*} $ consisting of 20 elements. For each $ v^{*} $ we partition its corresponding $ S^{*} $ into 5 sets. \textit{Left}: Histogram of the paired difference of the Mean Reciprocal Rank (MRR) between the ILP and Singleton. The effect size of the difference in MRR is quite large in favor of the ILP; Wilcoxon's signed rank test yields $ p \approx 0.0003 $.
    \textit{Right}: Recall @ $ k $ for various $ k $ for the ILP and three ``natural" competitors. We note that 16 is the maximum value and that the ILP convincingly outperforms the other methods.  
    }
    \label{fig:bing}
\end{figure}

The second real data example we consider is derived from Microsoft's Bing search navigation data collected over the course of 2019. The search navigation data consists of pairs of queries that are submitted to Bing in succession by a user within a browsing session. We consider the pairs of queries that correspond to a preidentified list of products in consumer electronics, household appliances, and gaming.

We use this data to construct a graph where the edge weight between two products in the graph is a normalized count of the pairs of queries that contain the two products. We remove self-loops and edges with edge weights below a user-selected threshold and analyze only the largest connected component, $ G_{Bing} $. $ G_{Bing} $ has $ n = |V| = 7,876 $ vertices and $ e = |E| = 46,062 $ directed weighted edges. 

We consider $ J = 100 $ different representations of $ G_{Bing} $. Each representation is a Node2Vec embedding \citep{grover2016node2vec} corresponding to different hyperparameter settings. 



Along with the representations, we are given three vertices of interest. All three vertices of interest correspond to phones. For each vertex of interest $ v^{*} $, we are given a set of vertices $ S^{*} $ known to be similar to $ v^{*} $. These sets were handpicked by a team of data scientists at Microsoft. Each $ S^{*} $ contains 20 products. A product was included either because it is a phone produced by the same company and is sufficiently close in generation or because it is of the same generation but produced by a different company.

To evaluate the proposed ILP we use the following scheme.
For each $(v^*,S^*)$ pair
we randomly partition $S^*$ ($|S^*|=20)$
into five subsets each of size four. These subsets are then each in turn used as $ S $ for the corresponding $ v^{*} $ and the ILP is evaluated on the remaining $|S^* \setminus S|=16$ vertices via MRR. This procedure resulted in a total of 15 different $ (v^{*}, S) $ pairs.

 

We compare the linear combination of the 100 representations found by the ILP to the nomination scheme that selects a ranker from the subset of the $ \{h^{d^{j}}\}_{j=1}^{100} $ that minimizes the objective function in \eqref{eq:main-prob}. That is, the second scheme uses the ranker induced by one of the $ 100 $ representations that minimizes the maximum rank of an element of $ S $. If two or more rankers tie then the ranker used is randomly selected from the argmin set. This scheme is referred to as ``Singleton" because it uses only a single representation.
Note that when a single representation optimizes \eqref{eq:main-prob} then the ILP and Singleton produce the same ranking.

The competing nomination schemes are, again, evaluated via MRR. The paired difference histogram and density estimate between the MRR from the ILP and the average MRR from Singleton are shown in Figure \ref{subfig:bing-mrr}. 
Testing the hypothesis ``$H_{0}$: Singleton is as good or better than ILP" results in $ p \approx 0.0003 $ from Wilcoxon's signed rank test.
(More to the point: ILP is {\it strictly} superior to Singleton for all 15 cases.)

\subsubsection*{Comparison to natural competitors}

The problem setting that we consider is, as far as we know, novel and it is (generally) unfair to compare our proposed ILP to methods developed for other settings \citep{hand2006classifier}. Instead, to get an understanding of the proposed method's utility, we compare the performance (as measured by Recall at $ k $ for various $ k $) of the ILP to Singleton and two other ``natural" algorithms: one based on Principal Component Analysis (``PCA") \citep{vidal2005generalized} and one based on random sampling (``Random"). 

In particular, for PCA we use the weight vector found by normalizing the absolute values of the eigenvalues of the first $ J =100 $ principal components of the personalized weight matrix. Thus there will be a different PCA solution for each query. We chose $ J = 100 $ so that the number of representations in the ILP solution and the PCA solution is the same. For Random, we randomly sample $ J=100 $ length unit vectors for the same amount of compute time as taken by the ILP and choose the best performing vector of weights, as measured by the objective function above \eqref{eq:main-prob}. The performances of the four algorithms (ILP, Singleton, PCA, Random) on the Bing experiment described above are shown in Figure \ref{subfig:bing-recall}. Notably, the ILP outperforms the natural competitors.

\subsection*{Drosophila} We consider a synaptic-resolution connectome of the \textit{Drosophila} larva brain (unpublished), including its learning and memory center (the mushroom body) 
\citep{eichler2017complete}. The connectome consists of $ n = |V| = 2965 $ neurons.
Edges are synapses -- directed. There are four edge types in the connectome: axon to dendrite $(|E_{AD}| = 54303)$, axon to axon $(|E_{AA}| = 34867)$, dendrite to dendrite $(|E_{DD}| = 10209)$, and dendrite to axon $ (|E_{DA}| = 4088)$. This connectome was manually annotated from electron microscopy imagery for a single \textit{Drosophila} larva brain \citep{ohyama2015multilevel, schneider2016quantitative}.

We consider $J=4$ different representations of the connectome obtained via spectral embedding of the individual Laplacian matrices corresponding to the four different directed weighted edge types. We omit embeddings corresponding to the spectral embedding of the adjacency matrix because of the relative sparsity of the dendrite to axon and dendrite to dendrite networks. 

These spectral embeddings yield a representation of each node for each edge type.
Note that since the connectome is directed, the left and right singular vectors differ. We use the concatenation of the two (embedding dimension $ m=11 $) as the representation, resulting in a $ 22 $ dimensional representation for each neuron for each edge type.  


The input neurons of the mushroom body (MBINs) are a well known and studied neuron type within the \textit{Drosophila} larva connectome \citep{saumweber2018functional}. For illustration purposes we consider each of the 26 MBINs in the brain as a $ v^{*} $ and 15 randomly selected MBINs as vertices known to be similar to $ v^{*} $. We evaluate the ILP and Singleton via MRR on the remaining 10 MBINs. Recall that Singleton uses a ranker from amongst the single representations that minimizes the objective function in \eqref{eq:main-prob}.

\begin{figure}[t]
    \centering
    \captionsetup[subfigure]{justification=centering}
    \begin{subfigure}{0.49\columnwidth}
        \includegraphics[width=\linewidth]{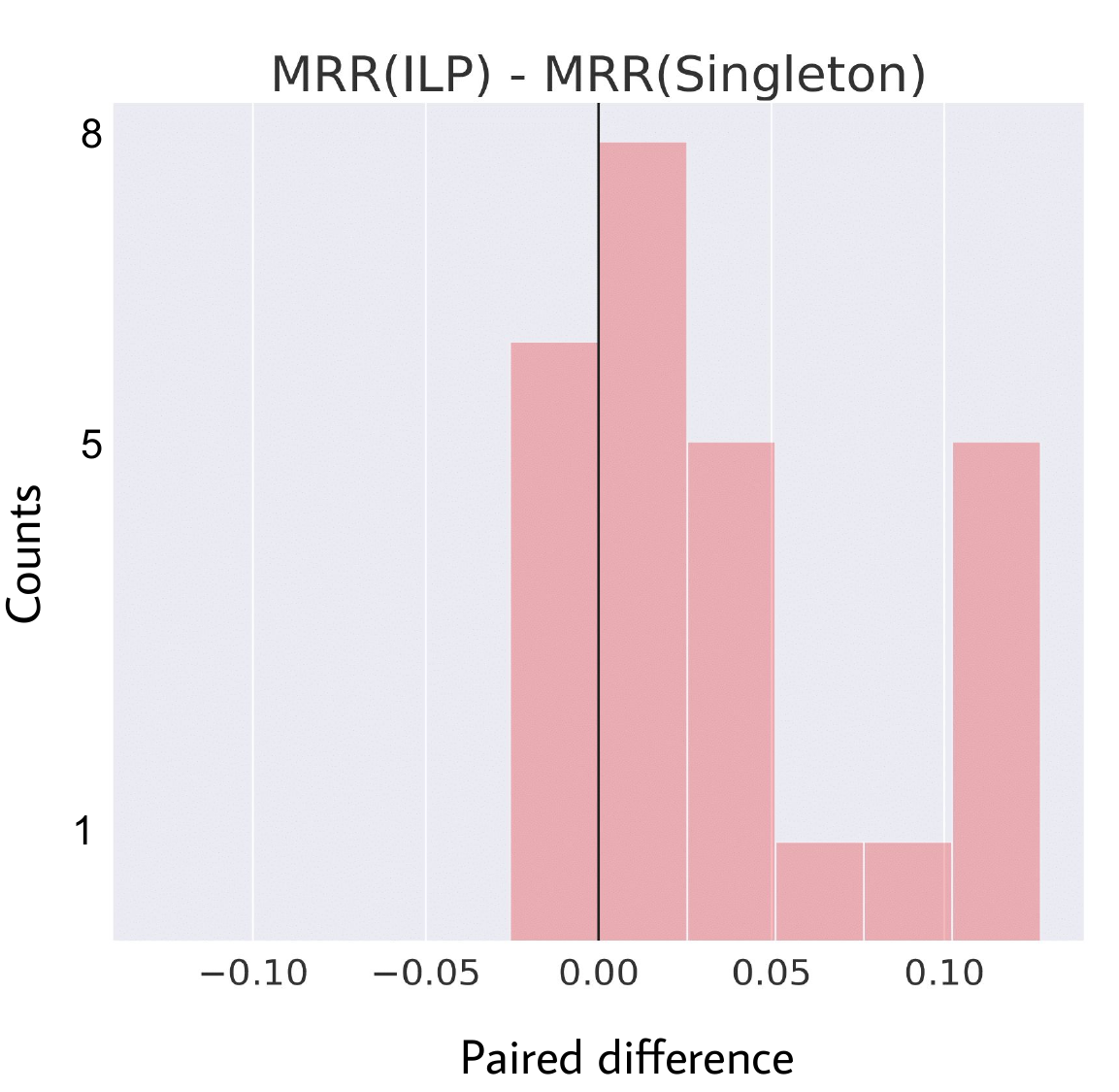}
        \caption{}
        \label{subfig:dros-singleton}
    \end{subfigure}
    \begin{subfigure}{0.49\columnwidth}
        \includegraphics[width=\linewidth]{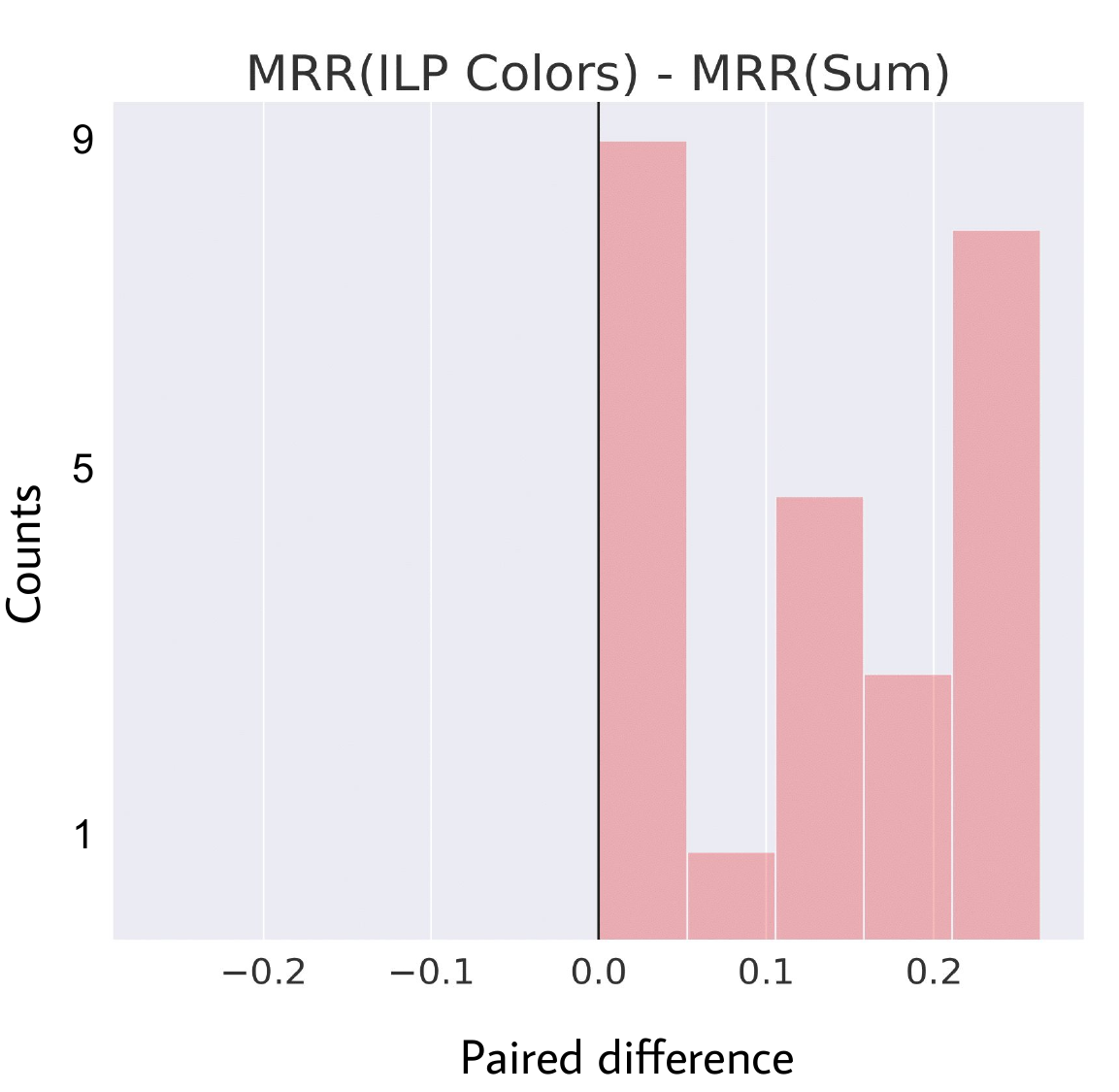}
        \caption{}
        \label{subfig:dros-sum}
    \end{subfigure}
    \caption{Each of the 26 MBINs in the connectome is considered as a $ v^{*} $. Of the remaining 25 MBINs, 15 are then chosen to be vertices known to be similar to $ v^{*} $. The nomination schemes are evaluated on the remaining 10 MBINs. \textit{Left}: The proposed ILP compared to Singleton. Wilcoxon's signed rank test yields $ p < 0.0001 $ in favor of the ILP. \textit{Right}: The proposed ILP compared to considering the sum of the four different edge types upstream from inference. The ``win" by the ILP demonstrates that the sum of the four edge types, which is what is typically considered for connectome-based inference, is insufficient (Wilcoxon's signed rank test yields $ p < 0.00001 $ in favor of the ILP).}
    \label{fig:drosophila}
\end{figure}

The paired difference histogram and estimated density between the MRR from the ILP and the MRR from Singleton are shown in Figure \ref{subfig:dros-singleton}. Testing the hypotheses ``$H_{0}$: Singleton is as good or better than ILP" results in $ p < 0.0001 $. This result indicates that the different edge types contain complementary information that, when put together, can yield superior inferential procedures as compared to procedures using only a single edge type. 

Of particular interest is the breakdown of solutions in terms of edge type.
As an example,
in one representative trial
the best singleton is AA (axon-to-axon),
while the ILP solution is the linear combination
of (AA, AD, DA, DD)
with weights (0.424, 0.123, 0, 0.453).
Indeed:
the singleton that minimizes the maximum rank of an element of $ S $ is always one of AA, AD, DD -- never DA,
and the ILP solution is always a linear combination of just these three edge types. 

\subsubsection*{Potential upstream impact}
When collecting synaptic data, there are two options: collect neuron-level data or collect \{axon, dendrite\}-level data. In effect, collecting neuron-level data is the same as collecting \{axon, dendrite\}-level data and then summing the counts across the different edge types or ``colors". 

The four different edge types possibly contain different and complementary information for inference related to neuron types in the connectome. This information is likely more useful than the single summary edge weight. Indeed, we demonstrate in Figure \ref{subfig:dros-sum} that considering the four different colors offers a significant improvement (Wilcoxon's signed rank test yields $ p < 0.00001$) over considering the sum of the edge types, for evaluating MBINs. Hence, if the inference task is to nominate MBINs, it is likely worthwhile to devote the extra resources to measure \{axon, dendrite\}-level synapses as opposed to the coarser-grained neuron-level synapses.

\section*{Conclusion}
We have presented an integer linear programming solution for learning to rank via combining representations.
This task is
of general interest, but for specificity we have presented our methodology and results from the perspective of vertex nomination in graphs.
The results presented herein -- analytic, simulation, and experimental -- demonstrate that this methodology is principled, practical, and effective.
Our methodology makes essentially no model assumptions;
just that we are given a query item,
we know the dissimilarity (from multiple perspectives) between it and a set of items to be ranked,
and we have a set of items known to be similar to the query.

The three experimental settings highlight three complementary aspects of our vertex nomination problem.
In the first -- dMRI -- the issue is how to utilize two different spectral embedding techniques (LSE and ASE) each known to uncover different graph structure \citep{Priebe5995,cape2019spectral}.
In the second -- Bing -- the issue is how to utilize a collection of pre-defined representations (Node2Vec, wherein optimal hyperparameter settings are unavailable at embedding-time, and in any event no one setting will be optimal for all tasks) for multiple post-embedding nomination tasks.
In the third -- {\it Drosophila} -- the issue is how to utilize multiple different graphs (in this case, synapse types) each on the same vertex set.
In all three settings, the ILP solution successfully optimizes our ``learning to rank'' objective function to obtain an effective ranking function.

Lastly, we note that though the ILP is a method for query-based information retrieval problems in general, in domains where strong representations are available out-of-the-box from pre-trained models, such as computer vision and natural language processing, the method will be less useful. Regardless, we think that the approach and extensions thereof will be both practical and effective for more nuanced domains.

\subsection*{Acknowledgements}
The authors thank 
Keith Levin,
Zachary Lubberts,
Ben Pedigo,
Anton Alyakin,
Eric Bridgeford,
Joshua Agterberg,
Jesus Arroyo, and
John Conroy
for constructive comments on an earlier draft of this manuscript.


\newpage



\vskip 0.2in
\bibliographystyle{elsarticle-num}

\bibliography{refs}

\newpage
\section*{Author biographies}

Hayden Helm is former research faculty at the Center for Imaging Sciences at Johns Hopkins University and former research intern at Microsoft Research. His research interests include statistical pattern recognition and modern machine learning. 

Amitabh Basu is an Associate Professor in the Applied Mathematics and Statistics department, with a secondary appointment in the Computer Science department, at Johns Hopkins University. His research interests lie in Optimization, geometry, Convex analysis, and the applications of these tools in Operations Research, Astronomy and Data Science.

Avanti Athreya was a visiting assistant professor with Duke University and a postdoctoral fellow with SAMSI prior to coming to Johns Hopkins University, in 2011, where she is currently an assistant research professor. Her research interests include statistical inference on random graphs and multiscale network analysis. 

Youngser Park holds joint appointments in the The Institute for Computational Medicine and the Human Language Technology Center of Excellence at Johns Hopkins University. His current research interests are clustering algorithms, pattern classification, and data mining for high-dimensional and graph data.

Joshua T. Vogelstein is an Assistant Professor in the Department of Biomedical Engineering, with joint appointments in Applied Mathematics and Statistics, Computer Science, Electrical and Computer Engineering, Neuroscience, and Biostatistics at Johns Hopkins University. 

Carey E. Priebe holds joint appointments at Johns Hopkins University in the Department of Computer Science, the Department of Electrical and Computer Engineering, and the Department of Biomedical Engineering, as well as the Center for Imaging Science, the Human Language Technology Center of Excellence, and the Mathematical Institute for Data Science. 

Michael Winding completed two Bachelor’s Degrees in Biology and Studio Art at the University of Notre Dame and a PhD in Cell Biology at Northwestern University. He is a postdoc working with Marta Zlatic, previously at HHMI Janelia and currently at the University of Cambridge, UK.

Marta Zlatic is a Croatian neuroscientist who is group leader at the MRC Laboratory of Molecular Biology in Cambridge, UK. Her research investigates how neural circuits generate behaviour.

Albert Cardona completed his undergraduate and graduate studies at the
University of Barcelona, postdoc at UCLA, started his lab at the
Institute of Neuroinformatics (Zurich), then at HHMI Janelia, and now is
a Programme Leader at the MRC LMB and associate professor in
neuroscience at the University of Cambridge, UK.

Patrick Bourke is a Software Engineer with Microsoft Research. He has extensive experience building and operating scalable services in industries such as retail, financial services and cloud computing. His work at MSR focuses on data engineering and software development for graph analysis and graph machine learning.

Jonathan Larson is a Principal Data Architect at Microsoft Research. His applied research work focuses on petabyte-scale data infrastructure, data science applications, network analytics, and information visualization.  He has applied experience in organizational science, neuroscience, cyber-security, counter-human trafficking, fraud analytics, mobile device analytics, media management, retail analytics, and real estate.

Marah I. Abdin is a Research Software Dev Engineer at Microsoft Research. She has been involved in multiple projects that use programmable hardware, nano-structural hardware, and machine learning. As she moves on in her career, her passion for problem-solving grows and she strives to continue exploring other engineering fields through progressive research.

Piali Choudhury is an Engineering Manager at Microsoft Research. She has been involved in deep systems level programming, networking technology, search, data mining, distributed storage, infrastructure and cloud services, precision medicine, real time communication platforms, integrative and applied ML/AI. Her passion is to empower research through engineering excellence.

Weiwei Yang is Principal SDE Manager at Microsoft Research. She is interested in resource efficient alt-SGD machine learning methods inspired by biological learning. The applied research group she leads aims to democratize AI by addressing issues of sustainability, robustness, scalability, and efficiency in ML.

Chris White is Managing Director, Microsoft Research Special Projects. He leads mission-oriented research and software development teams focusing on high risk problems.

\end{document}